\documentclass[a4paper,twoside]{article}

\usepackage{epsfig}
\usepackage{subcaption}
\usepackage{calc}
\usepackage{amssymb}
\usepackage{amstext}
\usepackage{amsmath}
\usepackage{amsthm}
\usepackage{multicol}
\usepackage{pslatex}
\usepackage{apalike}
\usepackage{SCITEPRESS}     

\usepackage{cite}
\usepackage{amsfonts}
\usepackage{algorithmic}
\usepackage{graphicx}
\usepackage{textcomp}
\usepackage{xcolor}
\usepackage{multirow}

\begin{document}

\newcommand\myfigure{%
\centering
\includegraphics[width=\textwidth]{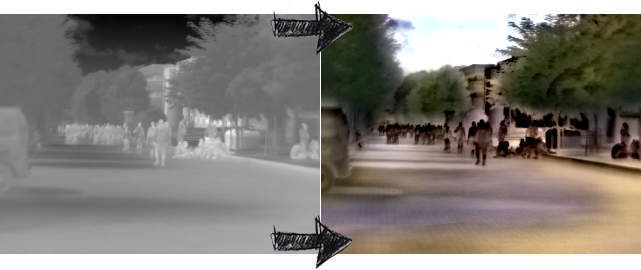}
\captionof{figure}{An example of mapping a thermal image to a color Visible image is presented. (left): a thermal image from the ULB17-VT.V2 test set, and, (right): its colorized counterpart. This approach generates color values consistent with the color Visible ground truth and preserves objects’ textures from the thermal representation.}
\label{fig:teaser}
}

\title{Robust Perceptual Night Vision in Thermal Colorization}

\author{
\authorname{Feras Almasri \sup{1}, Olivier Debeir \sup{1}}
\affiliation{Dept.LISA - Laboratory of Image Synthesis and Analysis, \\
 Université Libre de Bruxelles\\
 CPI 165/57, Avenue Franklin Roosevelt 50, 1050 Brussels, Belgium}
\email{falmasri@ulb.ac.be, odebeir@ulb.ac.be}
\myfigure{}
}


\abstract{Transforming a thermal infrared image into a robust perceptual colour Visible image is an ill-posed problem due to the differences in their spectral domains and in the objects' representations. Objects appear in one spectrum but not necessarily in the other, and the thermal signature of a single object may have different colours in its Visible representation. This makes a direct mapping from thermal to Visible images impossible and necessitates a solution that preserves texture captured in the thermal spectrum while predicting the possible colour for certain objects. In this work, a deep learning method to map the thermal signature from the thermal image’s spectrum to a Visible representation in their low-frequency space is proposed. A pan-sharpening method is then used to merge the predicted low-frequency representation with the high-frequency representation extracted from the thermal image. The proposed model generates colour values consistent with the Visible ground truth when the object does not vary much in its appearance and generates averaged grey values in other cases. The proposed method shows robust perceptual night vision images in preserving the object's appearance and image context compared with the existing state-of-the-art.}

\onecolumn \maketitle \normalsize \setcounter{footnote}{0} \vfill

\section{\uppercase{Introduction}}
\label{sec:introduction}

Humans have reasonable night vision with poor capabilities given improper environments. They have poor vision in low light conditions but with the advantage of rich colour vision in better lighting conditions. Human eyes have cone photoreceptor cells which are colour perception sensitive and rod photoreceptor cells which are receptive to brightness. The cones are unable to adapt well in low lighting conditions. 

\newpage

Colour vision is very important to the human brain. It helps to identify objects and to understand the surrounding environment. Studies \cite{cavanillas1999role} \cite{sampson1996assessment} have shown that human brain interpretation with colour vision improves the accuracy and the speed of object detection and recognition as compared to monochrome or false-colour visions. Due to this biologically limited interpretability, artificial night vision has become increasingly important in military missions, pharmaceutical studies, driving in darkness, and in security systems.

The use of thermal infrared cameras has seen an important increase in many applications, due to their long wavelength which allows capturing the objects invisible heat radiation despite lighting conditions. They are robust against some obstacles and illumination variations and can capture objects in total darkness. However, the human visual interpretability of thermal infrared images is limited, and so transforming thermal infrared images to Visible spectrum images is extremely important.   

The mapping process from monochrome Visible images into colour images is called colorization,  which has been broadly investigated in computer vision and image processing \cite{isola2017image} \cite{zhang2016colorful} \cite{larsson2016learning} \cite{guadarrama2017pixcolor}. However, it is an ill-posed problem because the two images are not directly correlated. A single object in the grayscale domain has a single representation while it might have different possible colour values in its true colour image counterpart. This is also true in the thermal images with additional challenging problems. For instance, a single object with different temperature conditions will have different thermal signature that can correspond to a single-colour value, while the thermal signature of two identical material objects at the same temperature conditions will look identical in the thermal infrared images, but have different colour values in their Visible image counterpart. 

Transforming thermal infrared images to Visible images is a very challenging task since they do not have the same electromagnetic spectrums  and so their representations are different. In grayscale image colorization, the problem is to transform the luminance values into only the chrominance values, while in thermal image colorization, the problem requires estimating the luminance and the chrominance given only the thermal signature. Accordingly, a delivered solution should consider all of these challenges and also provide a method for preserving the representation of the objects in the thermal spectrum, while predicting the possible colour of known relatively fixed in space and time objects, such as the sky, tree leaves, street, traffic signs.

This paper addresses the problem of transforming the thermal images to consistent perceptual Visible images using deep learning models. Our method predicts the low-frequency information of the Visible spectrum images and preserves the high-frequency information from the thermal infrared images. A pan-sharpening method is then used to merge these two bands and creates a plausible Visible image.

\section{\uppercase{Related Works}}

Earlier grayscale image colorization required human guidance to manually apply colour strokes to a selected region or to give a reference image with the same colour palette. This should help the model to assume the similar neighborhood intensity values and assign them a similar color, e.g. Scribble \cite{levin2004colorization}, or Similar images \cite{welsh2002transferring}, \cite{ironi2005colorization}. Recently, the successful applications of convolutional neural networks (ConvNet) have encouraged researchers to investigate automatic end-to-end ConvNet based model on the grayscale colorization problem \cite{cao2017unsupervised}, \cite{iizuka2016let}, \cite{cheng2015deep}, \cite{guadarrama2017pixcolor}.

A few researchers have investigated the colorization of near-infrared images (NIR) \cite{zhang2018tv}, \cite{limmer2016infrared} and have shown a high performance, due to the high correlation between the NIR and RGB images. Their two wavelengths differ only slightly in the red spectrum and thus they have similar Visible light representation correlated in the red channel. In contrast, thermal images taken from the long-wavelength infrared spectrum (LWI) do not correlate with the Visible images since they are measured by the emitted radiation linked to the objects' temperature. Therefore, predicting the colour of an object in its thermal signature requires a local and global understanding of the image context. 

Recently Berg et al. \cite{berg2018generating} and Nyberg et al. \cite{nyberg2018transforming} presented a fully automatic ConvNet on a thermal infrared to RGB image colorization problem using different objective functions. Their models illustrated a robust method against image pair misalignment. However, the generated images suffer from a high blur effect and artefacts in different locations in the images, e.g. missing objects from the scene, object deformations and some failure images. Kuang et al. in \cite{kuang2018thermal} used a conditional generative adversarial loss to generate a realistic Visible image, with the perceptual loss based on the VGG-16 model, the TV loss to ensure spatial smoothness, and the MSE as content loss. Their work presented better realistic colour representations with fine details but also suffered from the same artefacts, missing objects and object deformations.

The previous works were trained on the KAIST-MS dataset \cite{hwang2015multispectral} which consists of 95,000 thermal-Visible images captured from a device mounted on a moving vehicle. Images were captured during day and night by a thermal camera with an output size of 320x256 and interpolated to have the same size as the Visible images (640x512) using an unknown method and normalized using an unknown histogram equalization method. The procedure used to train the models in previous work reduces the size of the thermal images to their original size and then trains the models only on day time images. The frames were extracted from the video sequence, so it should be considered that, several subsequent images are very similar in most of the sets and  it is possible to overfit the dataset. It is also possible that the equalization coupled with the rescaling methods changed the thermal value distribution. Therefore, the proposed model is also trained on the ULB17-VT dataset \cite{almasri2018multimodal} which contains raw thermal images.

\section{\uppercase{Method}}

For this work, the target is to transform the thermal infrared images from their temperature representations to colour images. For this reason, this work builds on existing works that have looked at the thermal colorization problem and uses the proposed network architecture by Berg et al. \cite{berg2018generating} with small modifications adapted to our outputs. 

Preprocessing steps are assumed necessary when the ULB17-VT dataset is used. Images are normalized to $[0-1]$ using instance normalization in contrast with the KAIST-MS dataset which used histogram equalization. Spikes that occur with sharp low/high temperatures are detected and smoothed using a convolution kernel. 

The method proposed here is to transform the thermal image to low-frequency (LF) information in the colour Visible image space in a match with the LF information in the ground truth Visible image. The final colourized image is acquired by applying a post-processing pansharpening step. This process is done by merging the predicted Visible LF information with the high-frequency (HF) information extracted from the input thermal image. This step is assumed necessary to maintain an object's appearance from the thermal signature and to preserve it in the predicted colourized images. It also helps avoid high artefact occurrences when object representations are different between the two spectrums. 

\begin{figure}[ht]
\centerline{\includegraphics[scale=.34]{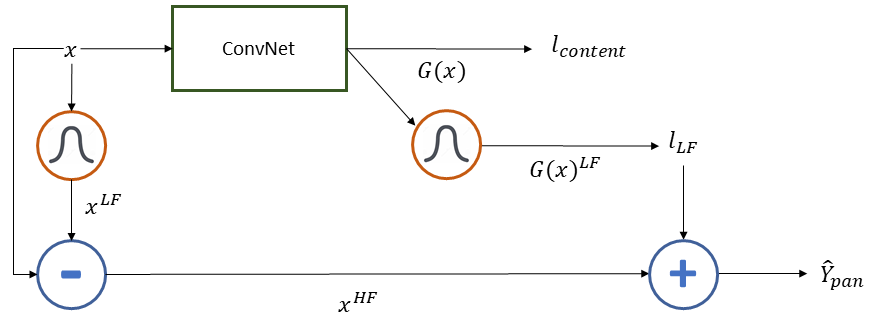}}
\caption{Proposed Model. Model (G) in orange is the Gaussian layer.}
\label{fig:Modelabs}
\end{figure}

\subsection{Proposed Model}

The proposed model, as illustrated in Fig.~\ref{fig:Modelabs}, takes the thermal image as input and generates a fully colourized Visible image. For this generated output, L1 content loss $l_{content}$ is used as an objective function to measure the dissimilarities with the ground truth Visible image. The low-frequency information is then obtained from the generated colourized image $G(x)^{LF}$ and from the ground truth Visible image $Y^{LF}$ by applying a Gaussian convolution layer with a kernel of width $25$ x $25$ and $\sigma = 12$. The dissimilarities between the LF information of the two images  is measured using the objective function $l_{lf}$ which is the MSE loss. The total loss is a weighted sum of the L1 and MSE multiplied by $\alpha = 10$ since the MSE loss value is smaller than L1.

\begin{equation}
 l_{total} = l_{content} +  \alpha \cdot l_{lf} 
  \label{eq:1}
\end{equation}

\subsection{Representation and Pre-Processing}

The pansharpening method is used as shown in Fig.~\ref{fig:Modelabs} as a final post-processing step. The thermal low-frequency information $x^{LF}$ is first obtained by applying a Gaussian layer on $x$. The thermal high-frequency information $x^{HF}$ is then extracted by subtracting $x^{LF}$ from $x$. The thermal image is represented with three channels in order to add them to the Visible RGB images. The final colourized thermal image $\hat{Y}_{pan}$ is obtained by adding the input $x^{HF}$ weighted by $\lambda$ to the generated low-frequency information $G(x)^{LF}$ as:
 
\begin{equation}
 \hat{Y}_{pan} = G(x)^{LF} +  \lambda x^{HF} 
  \label{eq:2}
\end{equation}

\begin{figure}[!ht]
\includegraphics[width=0.47\textwidth]{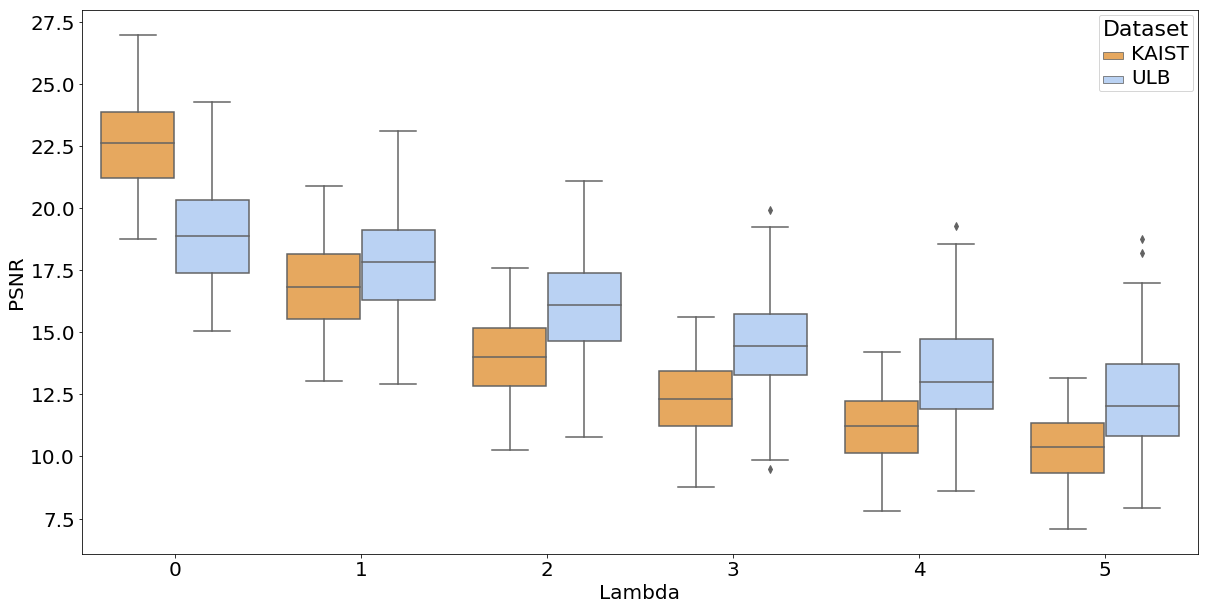}
\caption{Boxplot of PSNR for $\lambda = 0, 1, 2, 3, 4, 5$ on ULB17-VT.V2 test set and on KAIST-MS set00-V000 set}
\label{fig:boxplot}
\end{figure}

\begin{figure}[!ht]
    \centering
    
    \begin{subfigure}[b]{0.23\textwidth}
        \includegraphics[width=\textwidth]{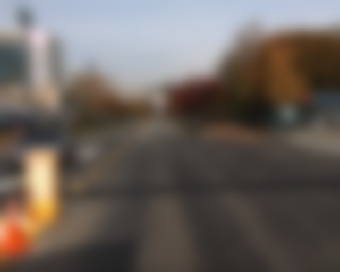}
        \caption{$\lambda = 0$}
    \end{subfigure}  
     \begin{subfigure}[b]{0.23\textwidth}
        \includegraphics[width=\textwidth]{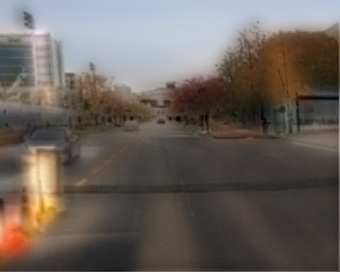}
        \caption{$\lambda = 1$}
    \end{subfigure}  \\
    \vspace{2pt}
    
    \begin{subfigure}[b]{0.23\textwidth}
        \includegraphics[width=\textwidth]{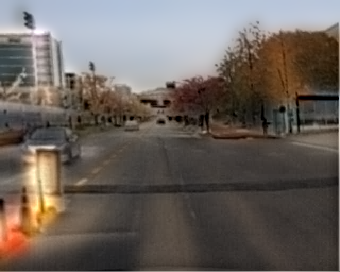}
        \caption{$\lambda = 2$}
    \end{subfigure}  
     \begin{subfigure}[b]{0.23\textwidth}
        \includegraphics[width=\textwidth]{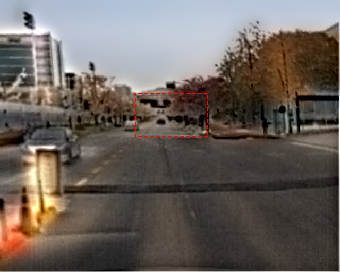}
        \caption{$\lambda = 3$}
    \end{subfigure}
    
    \caption{Pansharpening visualization from KAIST-MS dataset on S6V0I00000 with $\lambda = 0, 1, 2, 3$.}
    \label{fig:pansharpped}
\end{figure}

The pansharpening method is first applied on the ground truth Visible images to experience and visualize the pan-sharped colourized images before training the model. The thermal signature of the sky in the thermal images is very low with respect to other objects, while humans and other heated objects have a higher thermal signature. The normalization process makes the sky values very close to zero, while in the Visible images this value should be around one. For this reason, the thermal infrared images are inverted before any processing which results in a value around one for the sky in the thermal images. 

The proposed method relies on maintaining the high-frequency information taken from the thermal images, as this can reduce the evaluation results compared to the state-of-the-art when the pixel-wise measurement is used. For validation purposes,  the PSNR between $\hat{Y}_{pan}$ and $y$ with $\lambda = 0, 1, 2, 3, 4, 5$ was measured as shown in Fig.~\ref{fig:boxplot}. This gives an idea of the maximum validation value that can be achieved using the proposed model. The synthesised images are represented as a perceptual visualization quality as shown in Fig.~\ref{fig:pansharpped}. The value $\lambda = 3$ was chosen as a trade-off between better perceptual image quality and a reasonable PSNR with the average of $14.5$ for ULB17-VT.V2 and $12.31$ for KAIST-MS. If  $\lambda$ is decreased the PSNR value increases, but with less plausible perceptual images.

When the weighted thermal HF information is added to the Visible LF information, the synthesized image could have values out of the band $[0 - 1]$ in some areas. This results in a black or white color effect when the image is clipped to the range $[0 - 1]$ as shown in the red rectangle in Fig.~\ref{fig:pansharpped}. Re-normalizing the image instead of clipping can reduce the image contrast or affect the true colour values since the low frequency information on the three RGB channels is being obtained and added. This problem can be solved by exploring different normalization methods in the pre-processing step and different merging procedures in the post-processing step.

De-spiked thermal images are obtained using a convolution kernel of width $5$ x $5$, which replaces the centre pixel with the median value if the pixel value is three times greater than the standard deviation of the kernel area.

\subsection{Networks Architecture}

The network architecture proposed in \cite{berg2018generating} from their repository was used. \footnote{https://github.com/amandaberg/TIRcolorization}. Two models were trained as follows: 
\begin{itemize}
    \item TICPan-Bn  The proposed method using the network architecture in \cite{berg2018generating}. 

    \item TICPan  The proposed method using the same network architecture, and replacing the batch normalization layer with the instance normalization layer. It shows better enhancement in colour representations and in the metric evaluations.
\end{itemize}

\section{\uppercase{Experiments}}

\subsection{Dataset}
For this work  the ULB17-VT dataset \cite{almasri2018multimodal} which contains 404 Visible-thermal image pairs was used. The number of images was increased to 749 Visible-thermal images using the same device and 74 pairs were held for testing. Thermal images were extracted in their raw format and logged in with 16-bit float per-pixel. This new dataset, ULB-VT.v2, is available on \footnote{http://doi.org/10.5281/zenodo.3578267}. 

The KAIST-MS dataset \cite{hwang2015multispectral} was also used and the exiting works on thermal colorization problem were followed. Training was only done on day time images and resized the thermal images to their original resolution of $320$ x $256$ pixels. The images in KAIST-MS were recorded continuously during driving and stopping the car. This results in a high number of redundant images and explains the over-fitting behaviour and the failure results in previous work. For this reason, only every third image is taken in the training set to yield a set with 10,027 image pairs, while all of the images in the test set are used.

\subsection{Training Setup}
All experiments were implemented in Pytorch and performed on an NVIDIA TITAN XP graphics card. TIR2Lab \cite{berg2018generating} and TIC-CGAN  \cite{kuang2018thermal} were re-implemented and trained as explained in the original papers. 
 
The proposed model, TICPan, trained using ADAM optimizer with default Pytorch parameters and weights were initialized with He normal initialization \cite{he2015delving}. All experiments were trained for $1000$ epochs and the learning rate was initialized with $8e^{-4}$ with decay after 400 epochs. The LeakyReLU layers parameter was set to $\alpha=0.2$ and the dropout layer was set to $0.5$.

In each training batch, 32 cropped images of size $160$ x $160$ were randomly extracted. For each iteration, a random augmentation was applied by flipping horizontally or vertically and rotating in the $[-90^{\circ},90^{\circ}]$. Since the number of training images in KAIST-Ms is 14 times more than ULV-VT.v2, the number of iterations for the model to train on the ULV-VT.v2 was increased to match the model trained on KAIST-MS. 

For validation, the peak signal-to-noise ratio (PSNR), structural similarity (SSIM) and root-mean-square error (RMSE) were used between the generated colorized images and the true images.

\subsection{Quantitative Evaluation}

The proposed model was evaluated on transforming thermal infrared images to RGB images compared with the state-of-the-art using the measurement metrics shown in Table \ref{table:quanti}. 

The proposed model evaluation was performed on the full colorized thermal image, which is the result of the fusion of the predicted Visible LF information and the input thermal HF information. This resulted in a higher pixel-wise error compared to other models since the HF content of the image was taken from the thermal domain. However, our method achieved comparable results with the synthesized images as shown in Fig.~\ref{fig:boxplot}. 

It is believed that the pixel-wise metrics are not suitable for the colorization problem where the perception of the image has an important role. The TIR2Lab achieved higher evaluation values while their generated images are uninterruptable. TIC-CGAN has 12.266 million parameters that explain the overfitting behaviour in its generated images. TICPan-BN was excluded because it has the lowest evaluation values and less comparable quality images.

\begin{table*}[ht]
\caption{Average evaluation results on 74 images in ULB-VT version 2 dataset and 29,179 images in KAIST-MS dataset.}
\label{table:quanti}
\begin{center}
\begin{tabular}{@{} p{2cm}|p{2cm}|p{3cm}|p{2cm}|p{2cm}|p{2cm} @{}}
\hline\noalign{\smallskip}
Model  & Parameters & Dataset  & PSNR  & SSIM & RMSE \\ 
\noalign{\smallskip}
\hline
\noalign{\smallskip}
\multirow{2}{*}{TIR2Lab}    & \multirow{2}{*}{1.46M} & ULB-VT.V2 & 14.404 & 0.335 & 0.194 \\
                            & & KAIST-MS & 14.090 & 0.565 & 0.204  \\ \hline
                            
\multirow{2}{*}{TIC-CGAN}   & \multirow{2}{*}{12.266M} & ULB-VT.V2 & 15.475 & 0.313 & 0.174 \\                             & & KAIST-MS & 16.010 & 0.552 & 0.165  \\ \hline

\multirow{2}{*}{TIC-Pan-BN}    & \multirow{2}{*}{1.46M} & ULB-VT.V2 & 12.559 & 0.215 & 0.239 \\ 
                            & & KAIST-MS & 12.944 & 0.373 & 0.228   \\ \hline
                            
\multirow{2}{*}{TIC-Pan}    & \multirow{2}{*}{1.46M} & ULB-VT.V2 & 13.078 & 0.228 & 0.226 \\ 
                            & & KAIST-MS & 13.922 & 0.404 & 0.205   \\ \hline
\end{tabular}
\end{center}
\end{table*}

\subsection{Qualitative evaluation}
\begin{figure*}[ht]
\captionsetup[subfigure]{labelformat=empty}
    
    \rotatebox[origin=l]{90}{\bfseries Thermal image\strut}
     \begin{subfigure}[b]{0.23\textwidth}
        \includegraphics[width=\textwidth]{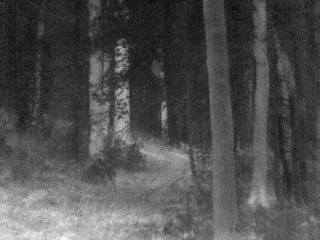}
    \end{subfigure} 
    \begin{subfigure}[b]{0.23\textwidth}
        \includegraphics[width=\textwidth]{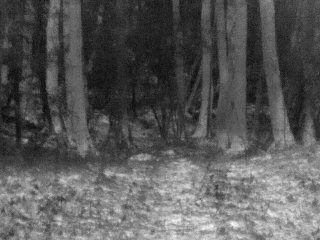}
    \end{subfigure}
    \begin{subfigure}[b]{0.23\textwidth}
        \includegraphics[width=\textwidth]{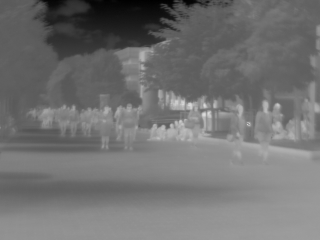}
    \end{subfigure}
    \begin{subfigure}[b]{0.23\textwidth}
        \includegraphics[width=\textwidth]{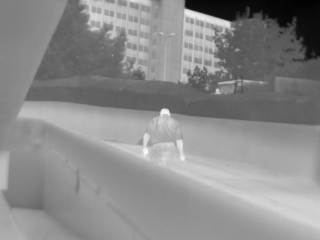}
    \end{subfigure}

    \rotatebox[origin=l]{90}{\bfseries TIR2Lab\strut}
     \begin{subfigure}[b]{0.23\textwidth}
        \includegraphics[width=\textwidth]{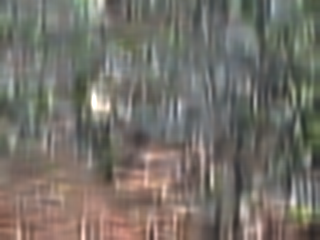}
    \end{subfigure} 
    \begin{subfigure}[b]{0.23\textwidth}
        \includegraphics[width=\textwidth]{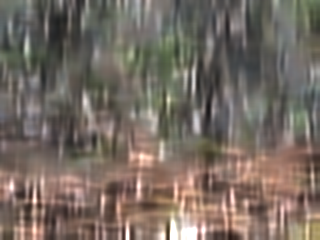}
    \end{subfigure}
    \begin{subfigure}[b]{0.23\textwidth}
        \includegraphics[width=\textwidth]{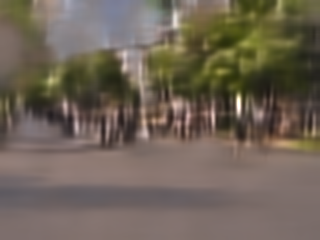}
    \end{subfigure}
    \begin{subfigure}[b]{0.23\textwidth}
        \includegraphics[width=\textwidth]{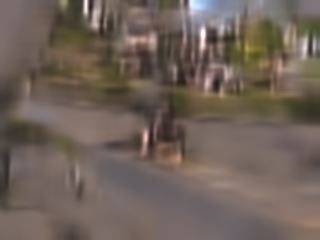}
    \end{subfigure}

    \rotatebox[origin=l]{90}{\bfseries TIC-CGAN\strut}
     \begin{subfigure}[b]{0.23\textwidth}
        \includegraphics[width=\textwidth]{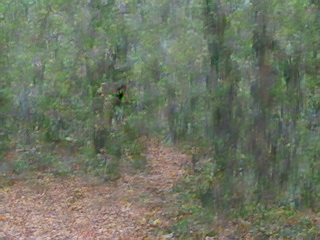}
    \end{subfigure} 
    \begin{subfigure}[b]{0.23\textwidth}
        \includegraphics[width=\textwidth]{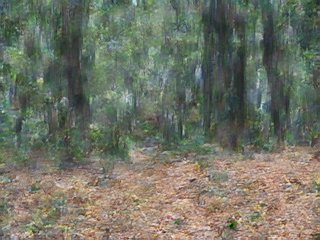}
    \end{subfigure}
    \begin{subfigure}[b]{0.23\textwidth}
        \includegraphics[width=\textwidth]{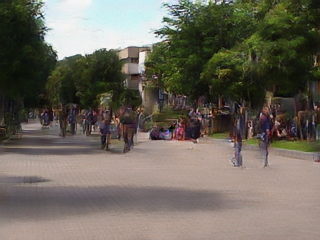}
    \end{subfigure}
    \begin{subfigure}[b]{0.23\textwidth}
        \includegraphics[width=\textwidth]{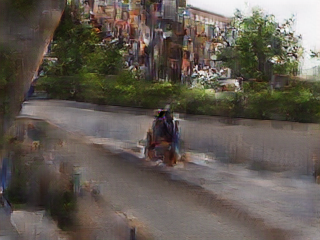}
    \end{subfigure}

    \rotatebox[origin=l]{90}{\bfseries TICPan\strut}
     \begin{subfigure}[b]{0.23\textwidth}
        \includegraphics[width=\textwidth]{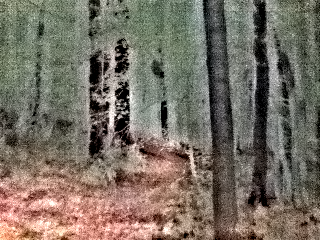}
    \end{subfigure} 
    \begin{subfigure}[b]{0.23\textwidth}
        \includegraphics[width=\textwidth]{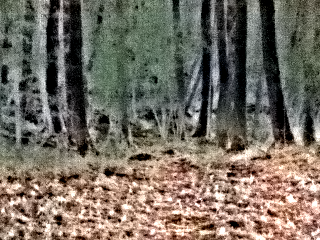}
    \end{subfigure}
    \begin{subfigure}[b]{0.23\textwidth}
        \includegraphics[width=\textwidth]{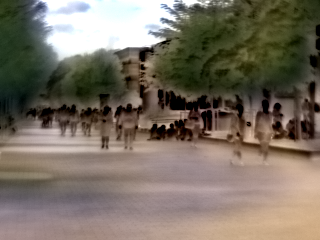}
    \end{subfigure}
    \begin{subfigure}[b]{0.23\textwidth}
        \includegraphics[width=\textwidth]{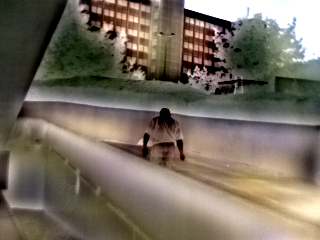}
    \end{subfigure}

    \rotatebox[origin=l]{90}{\hspace{10pt} \bfseries Visual image\strut}
    \begin{subfigure}[b]{0.23\textwidth}
        \includegraphics[width=\textwidth]{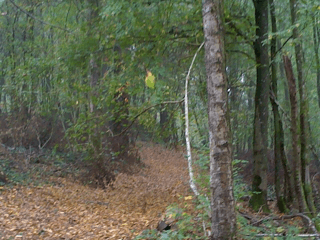}
         \caption{1508}
    \end{subfigure} 
    \begin{subfigure}[b]{0.23\textwidth}
        \includegraphics[width=\textwidth]{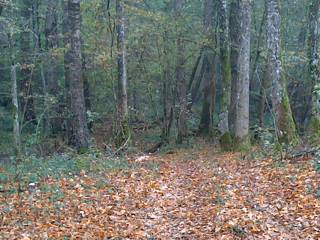}
         \caption{1476}
    \end{subfigure}
    \begin{subfigure}[b]{0.23\textwidth}
        \includegraphics[width=\textwidth]{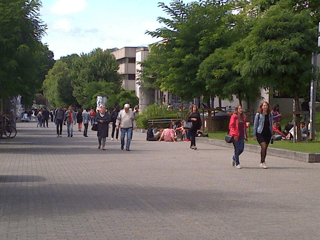}
         \caption{650}
    \end{subfigure}
    \begin{subfigure}[b]{0.23\textwidth}
        \includegraphics[width=\textwidth]{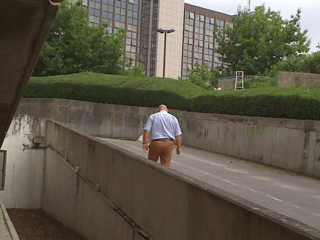}
         \caption{248}
    \end{subfigure}

    \caption{Examples of colorized results on ULB-VT.v2 test set. The numbers represent the image names.}
    \label{fig:qualiULB}
\end{figure*}

Four examples are presented in Fig.~\ref{fig:qualiULB} on the ULB17-VT.v2 dataset. The TIR2Lab model generated approximated good colour representations for trees with blur effect but failed to produce fine textures and to preserve the image content. On the hand, the TIC-CGAN model generated better image colour quality with fine textures and were more realistic. This is very recognizable, as an over-fitting behaviour, when the test image comes from the same distribution as the densely represented images in the training set such as image number (650).

TICPan generates images that have strong true colour values for objects that are relatively fixed in space and time, such as sky, tree leaves, and streets and buildings. Sky is represented in white or light blue colour, trees are in different shades of green, and streets and buildings also represented with approximated true colour values. However, objects like humans are represented in grey or in black due to the clipping effect. Our method assures that the object thermal signature does not disappear in image transformation or get deformed. The model cannot predict true colour values for the varying objects but it predicts an averaged colour value represented in grey and the final pansharpening process maintains their appearance in the generated colourized images.

In Fig.~\ref{fig:kaist} four examples are presented on the KAIST-MS dataset. The TIR2Lab method produced approximate good true chrominance values but it has heavily blurred images and suffers from recovering fine textures accurately. The produced artefacts are very obvious in the generated images and some objects, such as the walking person in (S6V3I03016) are missing in their outputs. The TIC-CGAN model produced better perceptual colourized thermal images with realistic textures and fine details, but they suffer from the same countereffects of missing objects and objects deformation. This is due to the use of GAN adversarial loss which learns the dataset distribution and estimates what should appear in each location, and also because of the large size of the model and its over-fitting behaviour. This is seen in (S8V2I01723) in the falsely generated road surface markings and in the missing person in (S6V3I03016). In contrast, the proposed TICPan model does not generate very plausible colour values in the KAIST-MS dataset but it generates robust perceptual night vision images that maintain objects' appearances.

\subsubsection{Deformation and missing Objects}

\begin{figure*}[ht!]
\captionsetup[subfigure]{labelformat=empty}
    
    \rotatebox[origin=l]{90}{\bfseries Thermal image\strut}
     \begin{subfigure}[b]{0.23\textwidth}
        \includegraphics[width=\textwidth]{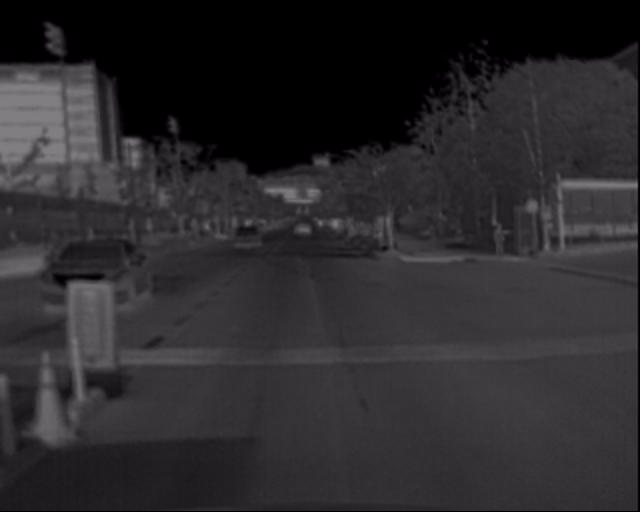}
    \end{subfigure}
    \begin{subfigure}[b]{0.23\textwidth}
        \includegraphics[width=\textwidth]{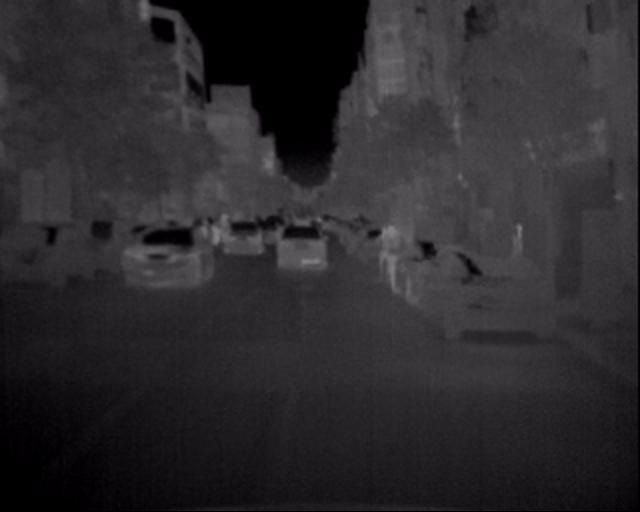}
    \end{subfigure}
    \begin{subfigure}[b]{0.23\textwidth}
        \includegraphics[width=\textwidth]{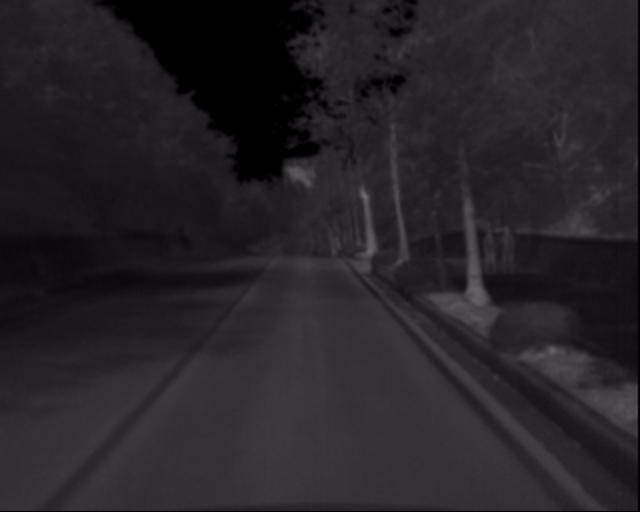}
    \end{subfigure}
     \begin{subfigure}[b]{0.23\textwidth}
        \includegraphics[width=\textwidth]{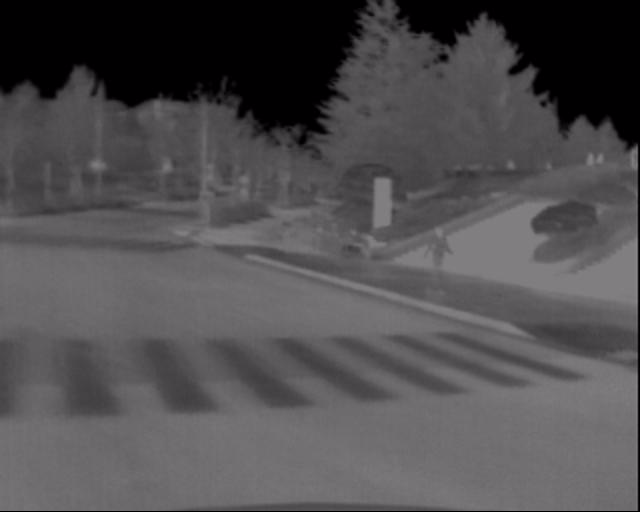}
    \end{subfigure}

    \rotatebox[origin=l]{90}{\bfseries TIR2Lab\strut}
     \begin{subfigure}[b]{0.23\textwidth}
        \includegraphics[width=\textwidth]{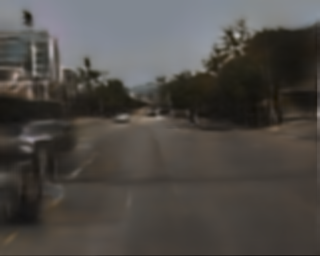}
    \end{subfigure}
    \begin{subfigure}[b]{0.23\textwidth}
        \includegraphics[width=\textwidth]{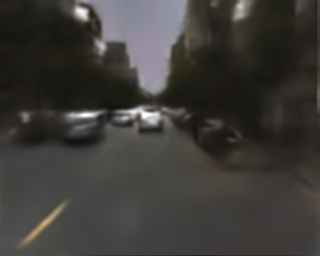}
    \end{subfigure}
    \begin{subfigure}[b]{0.23\textwidth}
        \includegraphics[width=\textwidth]{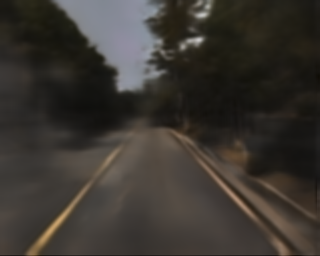}
    \end{subfigure}
     \begin{subfigure}[b]{0.23\textwidth}
        \includegraphics[width=\textwidth]{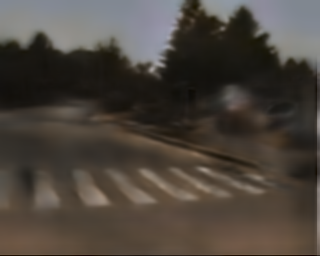}
    \end{subfigure}

    \rotatebox[origin=l]{90}{\bfseries TIC-CGAN\strut}
     \begin{subfigure}[b]{0.23\textwidth}
        \includegraphics[width=\textwidth]{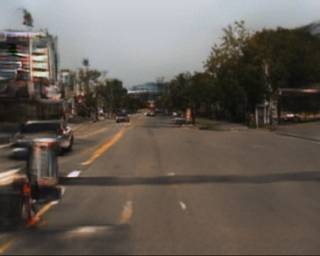}
    \end{subfigure}
    \begin{subfigure}[b]{0.23\textwidth}
        \includegraphics[width=\textwidth]{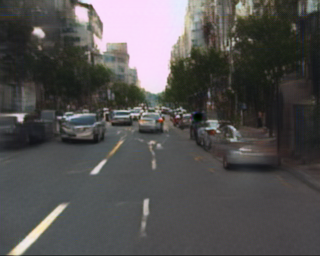}
    \end{subfigure}
    \begin{subfigure}[b]{0.23\textwidth}
        \includegraphics[width=\textwidth]{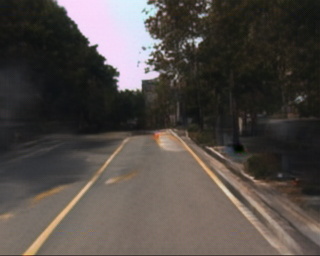}
    \end{subfigure}
     \begin{subfigure}[b]{0.23\textwidth}
        \includegraphics[width=\textwidth]{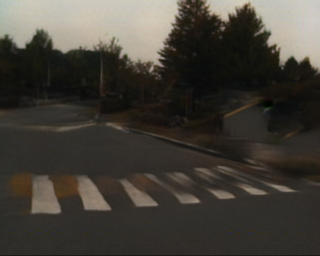}
    \end{subfigure} 
    
    \rotatebox[origin=l]{90}{\bfseries TICPan\strut}
     \begin{subfigure}[b]{0.23\textwidth}
        \includegraphics[width=\textwidth]{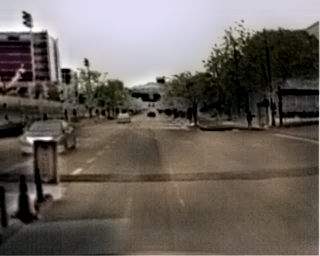}
    \end{subfigure}
    \begin{subfigure}[b]{0.23\textwidth}
        \includegraphics[width=\textwidth]{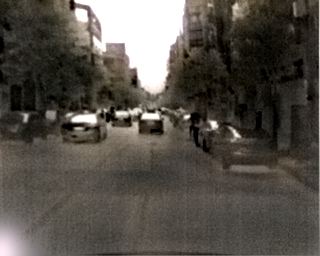}
    \end{subfigure}
    \begin{subfigure}[b]{0.23\textwidth}
        \includegraphics[width=\textwidth]{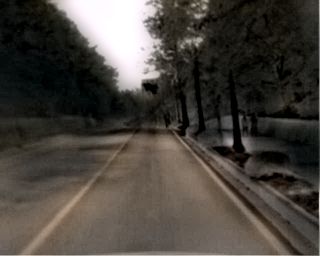}
    \end{subfigure}
     \begin{subfigure}[b]{0.23\textwidth}
        \includegraphics[width=\textwidth]{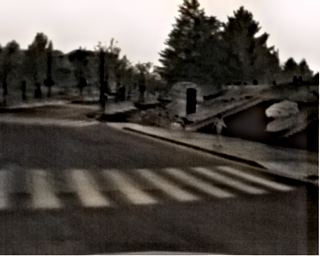}
    \end{subfigure} 
    
    \rotatebox[origin=l]{90}{\hspace{10pt} \bfseries Visual image\strut}
     \begin{subfigure}[b]{0.23\textwidth}
        \includegraphics[width=\textwidth]{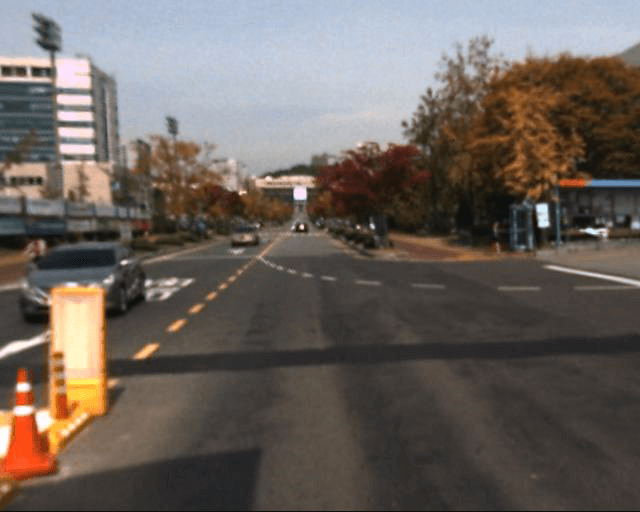}
        \caption{S6V0I00000}
    \end{subfigure}
    \begin{subfigure}[b]{0.23\textwidth}
        \includegraphics[width=\textwidth]{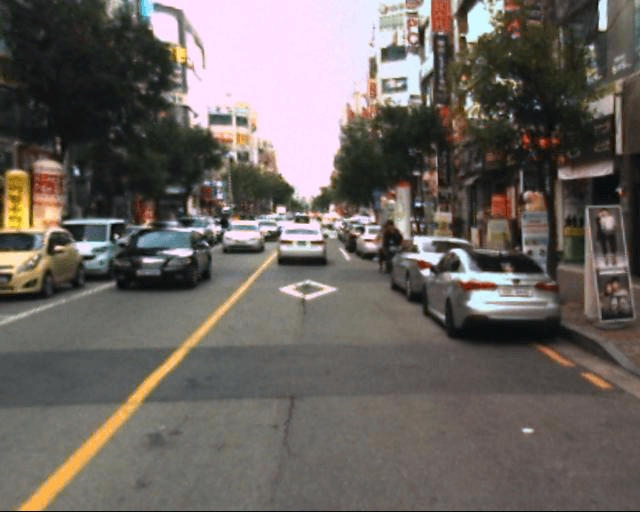}
        \caption{S8V2I01723}
    \end{subfigure}
    \begin{subfigure}[b]{0.23\textwidth}
        \includegraphics[width=\textwidth]{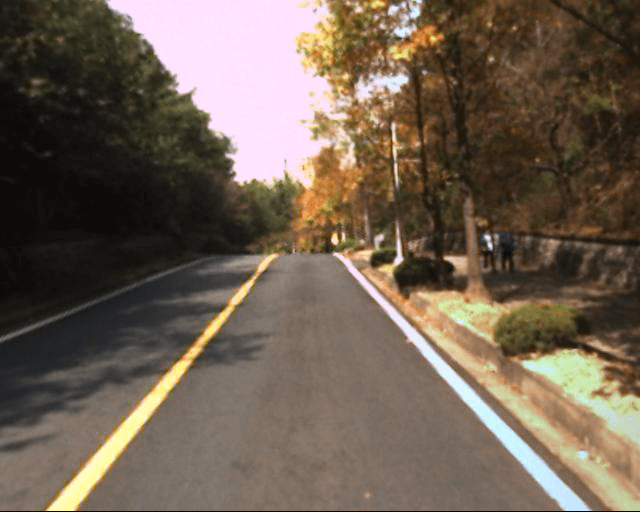}
        \caption{S6V2I00188}
    \end{subfigure}
     \begin{subfigure}[b]{0.23\textwidth}
        \includegraphics[width=\textwidth]{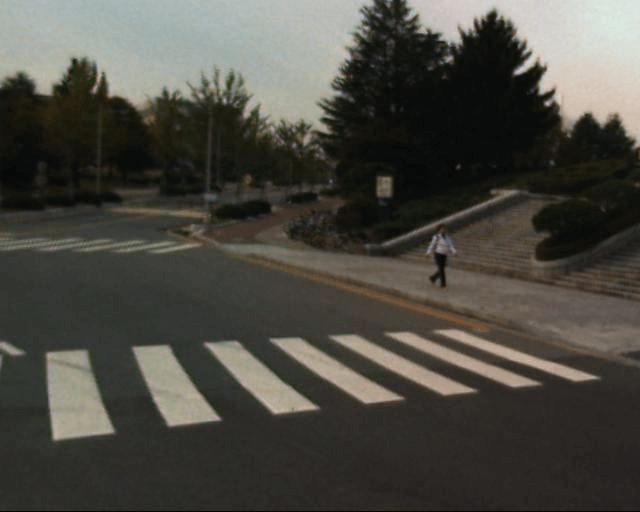}
        \caption{S6V3I03016}
    \end{subfigure} 
  
    \caption{Examples of colorized results on KAIST-MS test set. The numbers represent the image place and their names.}
    \label{fig:kaist}
\end{figure*}

Fig.~\ref{fig:kaist} shows missing objects in the TIC-CGAN generated images, such as the person in (S0V0I00601) and the cars in (S0V0I01335). We can also recognize the object deformation in  image number (428) and image number (598), while in the TICPan model objects are retained in the generated images. 

\begin{figure}[htpb]
    \centering
    
    \begin{subfigure}[b]{0.15\textwidth}
        \includegraphics[width=\textwidth]{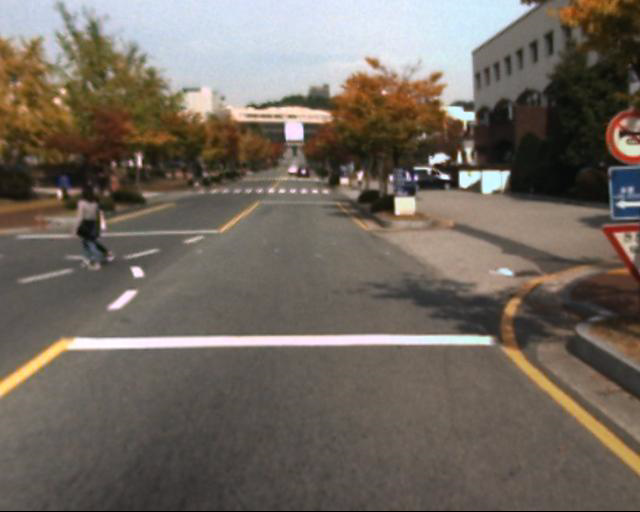}
    \end{subfigure}  
     \begin{subfigure}[b]{0.15\textwidth}
        \includegraphics[width=\textwidth]{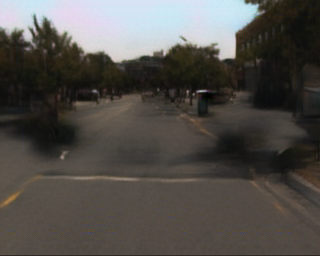}
    \end{subfigure}  
    \begin{subfigure}[b]{0.15\textwidth}
        \includegraphics[width=\textwidth]{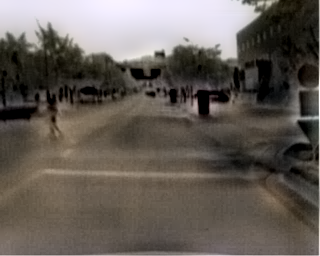}
    \end{subfigure} \\
    \vspace{2pt}
    
    \begin{subfigure}[b]{0.15\textwidth}
        \includegraphics[width=\textwidth]{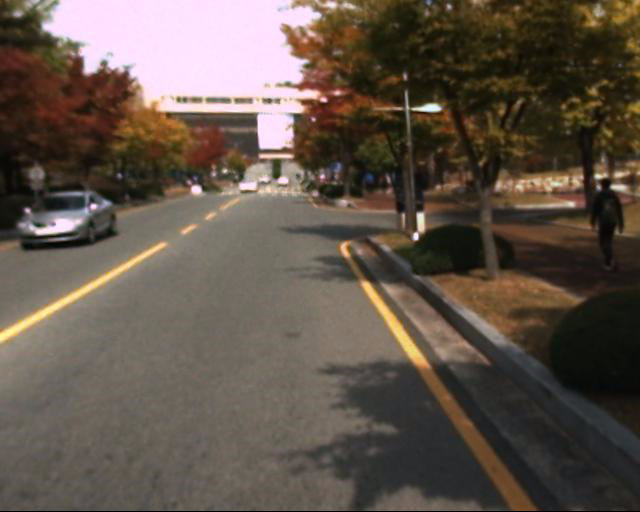}
    \end{subfigure}  
     \begin{subfigure}[b]{0.15\textwidth}
        \includegraphics[width=\textwidth]{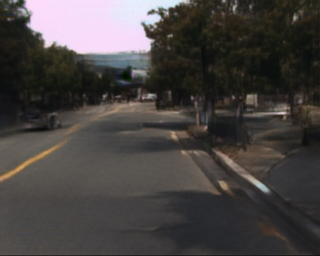}
    \end{subfigure}
    \begin{subfigure}[b]{0.15\textwidth}
        \includegraphics[width=\textwidth]{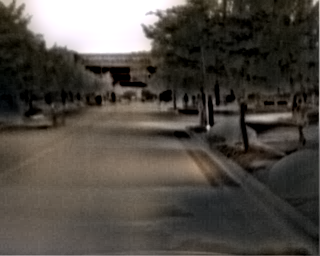}
    \end{subfigure} \\
    
    \begin{subfigure}[b]{0.15\textwidth}
    \includegraphics[width=\textwidth]{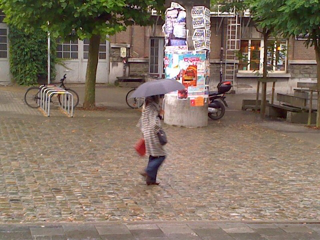}
    \end{subfigure}  
     \begin{subfigure}[b]{0.15\textwidth}
        \includegraphics[width=\textwidth]{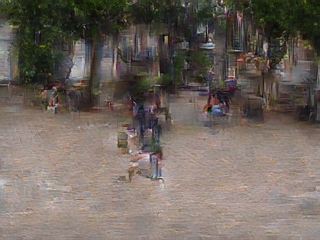}
    \end{subfigure}
    \begin{subfigure}[b]{0.15\textwidth}
        \includegraphics[width=\textwidth]{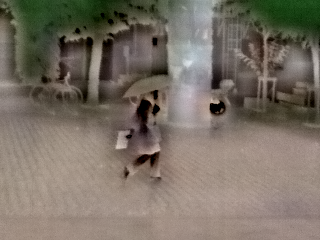}
    \end{subfigure} \\
    
    \begin{subfigure}[b]{0.15\textwidth}
    \includegraphics[width=\textwidth]{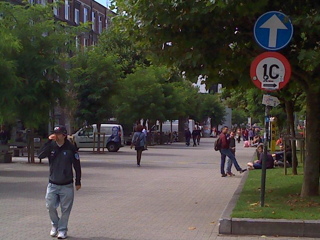}
    \end{subfigure}  
     \begin{subfigure}[b]{0.15\textwidth}
        \includegraphics[width=\textwidth]{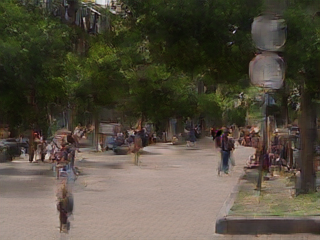}
    \end{subfigure}
    \begin{subfigure}[b]{0.15\textwidth}
        \includegraphics[width=\textwidth]{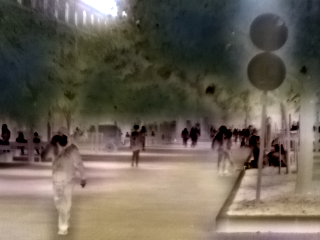}
    \end{subfigure} \\
    
    \caption{From left to right: True RGB, TIC-CGAN and TICPan. From top to bottom: (S0V0I00601) and (S0V0I01335) form KAIST-MS and (428) and (598) from ULB-VT.v2}
    \label{fig:defmis}
\end{figure}

\subsubsection{Overfitting behavior}

\begin{figure}[htpb]
    \centering
    
    \begin{subfigure}[b]{0.22\textwidth}
        \includegraphics[width=\textwidth]{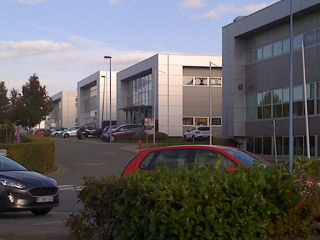}
    \end{subfigure}  
     \begin{subfigure}[b]{0.22\textwidth}
        \includegraphics[width=\textwidth]{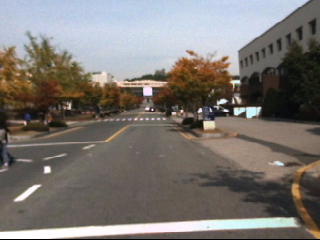}
    \end{subfigure}

    \begin{subfigure}[b]{0.22\textwidth}
        \includegraphics[width=\textwidth]{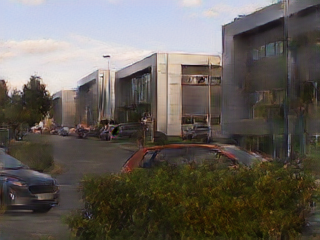}
    \end{subfigure} 
    \begin{subfigure}[b]{0.22\textwidth}
        \includegraphics[width=\textwidth]{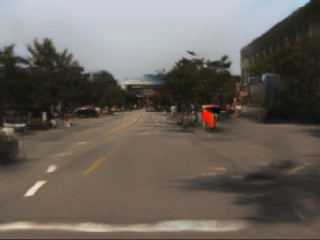}
    \end{subfigure}

    \begin{subfigure}[b]{0.22\textwidth}
        \includegraphics[width=\textwidth]{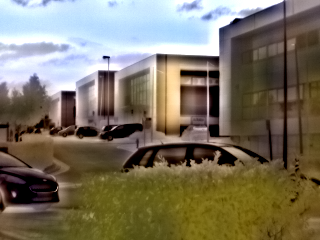}
    \end{subfigure}  
     \begin{subfigure}[b]{0.22\textwidth}
        \includegraphics[width=\textwidth]{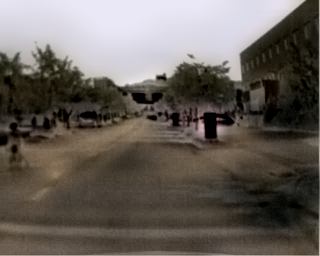}
    \end{subfigure}
    
    \caption{From left to right: (1250) from ULB-VT.v2 and (S0V0I00613) from KAIST-MS test set. From top to bottom: True RGB, TIC-CGAN and TICPan.}
    \label{fig:overfit}
\end{figure}

Fig.~\ref{fig:overfit} illustrates the over-fitting problem in the TIC-CGAN model. Because of its size, it has 12M parameters and is 12 times bigger than the proposed model. This makes it very easy for the model to overfit the dataset and not perform generalisation in the unseen data. In image number (1250), the model can predict the exact colour of the two cars because a similar image appeared in the training set. In the second image number (S0V0I00613), whenever an object comes from the left with a size similar to a bus, the model will predict it as a bus with red colour. The TICPan model cannot predict the exact colour of cars, but instead generates an average grey colour. 

\subsubsection{Night vision}

\begin{figure}[ht]
    \centering
    
    \begin{subfigure}[b]{0.23\textwidth}
        \includegraphics[width=\textwidth]{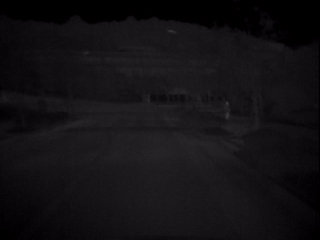}
    \end{subfigure}  
    \begin{subfigure}[b]{0.23\textwidth}
        \includegraphics[width=\textwidth]{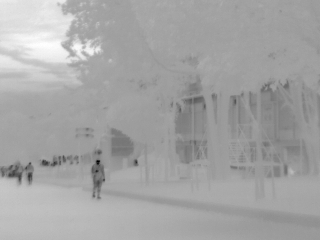}
    \end{subfigure}  
    
     \begin{subfigure}[b]{0.23\textwidth}
        \includegraphics[width=\textwidth]{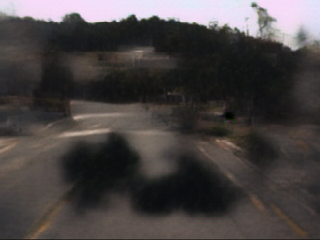}
    \end{subfigure}  
    \begin{subfigure}[b]{0.23\textwidth}
        \includegraphics[width=\textwidth]{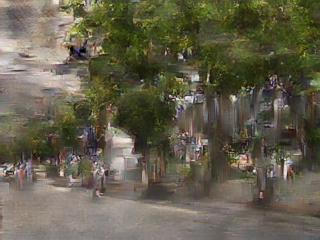}
    \end{subfigure} 
    
    \begin{subfigure}[b]{0.23\textwidth}
        \includegraphics[width=\textwidth]{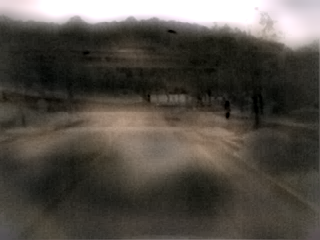}
    \end{subfigure} 
    \begin{subfigure}[b]{0.23\textwidth}
    \includegraphics[width=\textwidth]{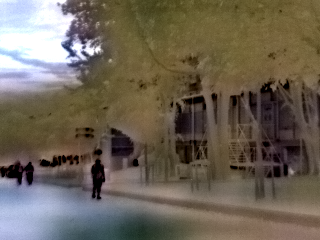}
    \end{subfigure} 
    
    \begin{subfigure}[b]{0.23\textwidth}
        \includegraphics[width=\textwidth]{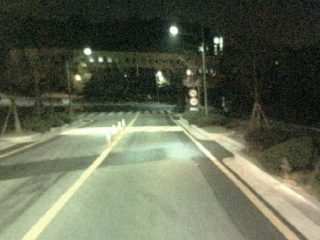}
    \end{subfigure} 
    \begin{subfigure}[b]{0.23\textwidth}
    \includegraphics[width=\textwidth]{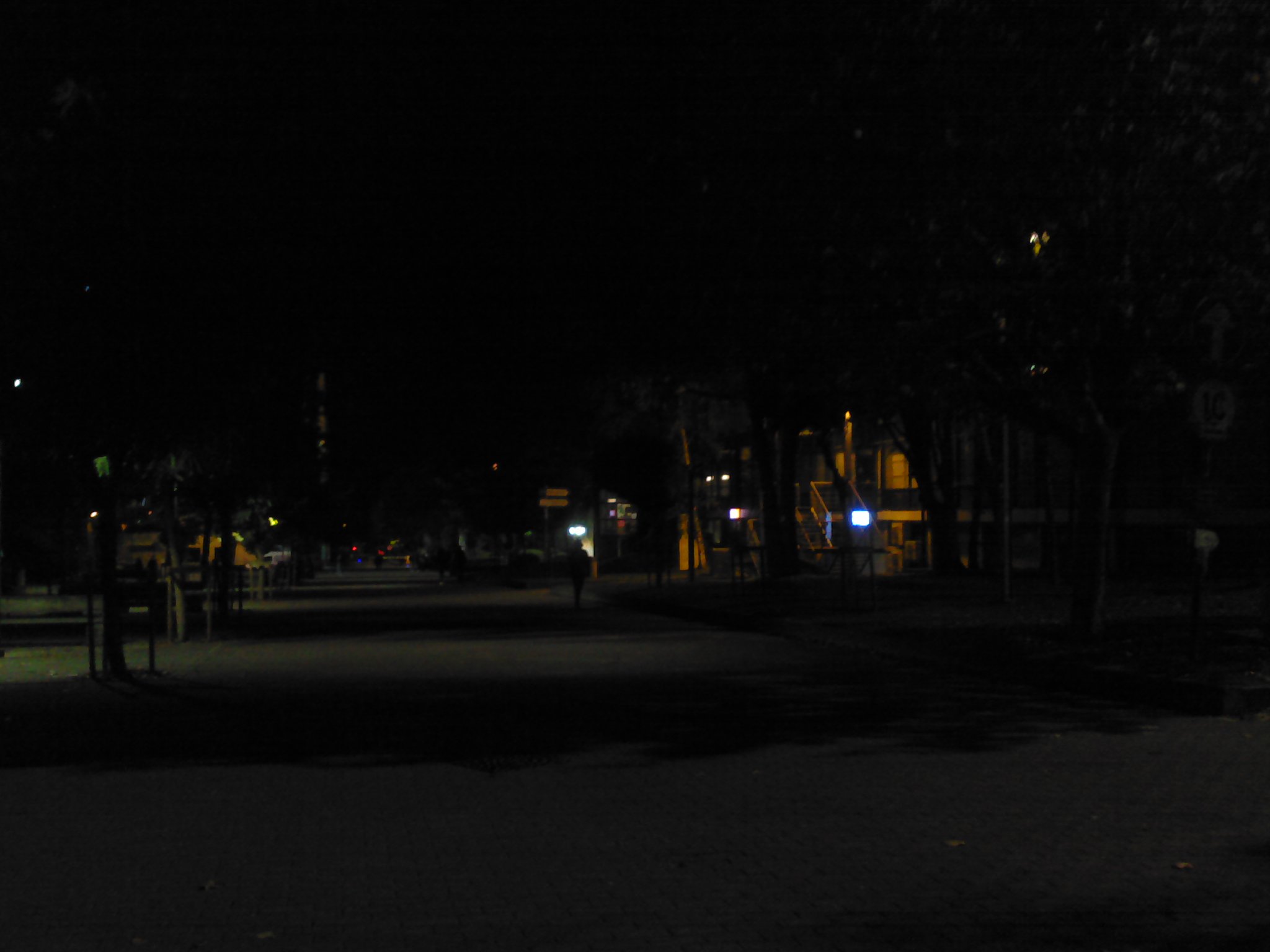}
    \end{subfigure} 

    \caption{From top to bottom: Thermal image, TIC-CGAN and TICPan. Left (S9V0I00000) from KAIST-MS and right (1784) ULB-VT.v2}
    \label{fig:night}
\end{figure}

The TIC-CGAN model failed to generate interpretable images using images that were taken at night, because the image distribution and the image contrast were different from the training images. However, the TICPan model does not suffer from this failure thanks to the pansharpenning process as shown in Fig.~\ref{fig:night}. In image number (1784), the true RGB image is completely dark and the TICPan model generates a robust perceptual night vision image as compared to the TIC-CGAN model. This is also illustrated in image number (S9V0I00000), where the TICPan model generates a night vision image with less artefacts than the TIC-CGAN model. It should be noted that these artefacts are due to the histogram equalization method used in KAIST-MS.

\section{\uppercase{Conclusions}}
\label{sec:conclusion}

The objective in this study was to address the problem of transforming thermal infrared images to Visible images with robust perceptual night vision quality. In contrast to the existing methods that map images automatically from their thermal signature to chrominance information, our proposed model seeks to maintain the appearance of objects in their thermal representation from the thermal images and to predict possible colour values. 

The evaluation showed that the proposed model has better perceptual images with fewer artefacts and the best representation for night images. This confirms the model generalization capability. The generated images are robust and reliable enabling users to better interpret the images while using night vision. For objects or cases in which missing or deformed objects can cause dramatic accidents, the pan sharpening process is of critical necessity. 


\section*{\uppercase{Acknowledgements}}
This work was supported by the European Regional Development Fund (ERDF) and the Brussels-Capital Region within the framework of the Operational Programme 2014-2020 through the ERDF-2020 project F11-08 ICITY-RDI.BRU.
We thank Thermal Focus BVBA for their support.

\bibliographystyle{apalike}
{\small
\bibliography{example}}



\end{document}